\documentclass[final,5p,times]{elsarticle}

\journal{arXiv}

\usepackage{hyperref}

\makeatletter
\g@addto@macro{\UrlBreaks}{\UrlOrds}
\makeatother

\usepackage{xcolor}

\usepackage{multirow}
\usepackage{booktabs}
\usepackage[para,online]{threeparttable}
\usepackage{float}
\usepackage{paralist}
\usepackage{tabularx}

\makeatletter \newif\if@restonecol \makeatother  
\usepackage{algorithm}
\usepackage{algorithmic}

\usepackage{booktabs} %
\usepackage{amsmath,amsfonts}
\usepackage{siunitx}

\usepackage{subfig}

\usepackage{epsfig}
\usepackage{epstopdf}
\usepackage{graphics}
\usepackage{graphicx}
\DeclareGraphicsExtensions{.eps,.gz,.eps,.pdf,.png,.jpg}

\graphicspath{{./}}

\newcommand{\sstitle}[1]{\smallskip\noindent\textbf{#1.\/}}

\DeclareMathOperator*{\argmin}{arg\,min}

\usepackage{pifont}%
\newcommand{\cmark}{\ding{51}}%
\newcommand{\xmark}{\ding{55}}%

\def\Snospace~{\S{}}

\makeatletter
\newcommand{\ALC@uniqueautorefname}{Line}
\makeatother

\makeatletter
\newcommand{\removelatexerror}{\let\@latex@error\@gobble}
\makeatother

\makeatletter
\def\input@path{{./}{../}}
\makeatother

\begin{document}

\setlength{\belowdisplayskip}{2pt}
\setlength{\belowdisplayshortskip}{2pt}
\setlength{\abovedisplayskip}{2pt}
\setlength{\abovedisplayshortskip}{2pt}

\begin{frontmatter}

\title{Model-Free Counterfactual Subset Selection at Scale}

\author{Minh Hieu Nguyen}
\author{Viet Hung Doan}
\author{Anh Tuan Nguyen}
\author{Jun Jo}
\author{Quoc Viet Hung Nguyen}

%\maketitle

\begin{abstract}
Ensuring transparency in AI decision-making requires interpretable explanations, particularly at the instance level. Counterfactual explanations are a powerful tool for this purpose, but existing techniques frequently depend on synthetic examples, introducing biases from unrealistic assumptions, flawed models, or skewed data. Many methods also assume full dataset availability, an impractical constraint in real-time environments where data flows continuously. In contrast, streaming explanations offer adaptive, real-time insights without requiring persistent storage of the entire dataset. This work introduces a scalable, model-free approach to selecting diverse and relevant counterfactual examples directly from observed data. Our algorithm operates efficiently in streaming settings, maintaining \(O(\log k)\) update complexity per item while ensuring high-quality counterfactual selection. Empirical evaluations on both real-world and synthetic datasets demonstrate superior performance over baseline methods, with robust behavior even under adversarial conditions.

%\keywords{
%streaming explanation 
%\and
%counterfactual selection
%\and 
%multi-class
%\and 
%case-based explanation
%\and 
%bias awareness
%}
\end{abstract}

\end{frontmatter}

\section{Introduction}
\label{sec:intro}

Explainable AI (XAI) is crucial for modern ML~\cite{dwivedi2023explainable}, driven by legal mandates like GDPR~\cite{wachter2017counterfactual} and ethical guidelines. Lack of explainability limits ML adoption in high-stakes areas, where biased models can lead to unfair outcomes such as credit denials or job rejections~\cite{barocas2020hidden}. Counterfactual explanations address this by showing changes needed to reverse outcomes, offering actionable feedback~\cite{schleich2021geco}, aligning with GDPR~\cite{wachter2017counterfactual}, and enabling users to explore ``what-if'' scenarios, while helping developers debug and monitor models effectively.

\begin{figure}[!h]
	\centering
%	\vspace{-2.5em}
	 \begin{minipage}{.5\linewidth}
    \centering
    \captionsetup{type=table}
     \caption{{Functionality comparison.}}
    \label{tab:functionality}
  \resizebox{1.0\linewidth}{!}{
    \begin{threeparttable}
      \begin{tabular}{l c c c c c c}
    \hline
     & DiCE~\cite{mothilal2020explaining} & MACE~\cite{karimi2020model} & GeCo~\cite{schleich2021geco} & CFPlan~\cite{BuiNN22} & FRPD~\cite{nguyen2023feasible} & \textbf{Ours} \\ 
    \hline
    Model-free & \xmark  & \xmark & \xmark & \xmark & \xmark & \cmark \\ 
    Reality & \xmark  & \xmark & \xmark & \xmark & \xmark & \cmark \\ 
    Streaming  & \xmark  & \xmark & \xmark & \xmark & \xmark & \cmark \\ 
    Multiplicity  & \cmark  & \xmark & \cmark  & \cmark & \cmark & \cmark\\ 
    Diversity  & \cmark  & \cmark & \xmark \tnote{1} & \xmark & \cmark & \cmark \\ 
 \hline
    \end{tabular}%
    \begin{tablenotes}
\item[1] Although GeCO used genetic algorithm to generate counterfactuals, which is claimed to be diverse, they did not measure it explicitly.
\end{tablenotes}
\end{threeparttable}
  }
      \end{minipage}
      \begin{minipage}{0.44\linewidth}
		\centering
	\includegraphics[width=0.6\linewidth]{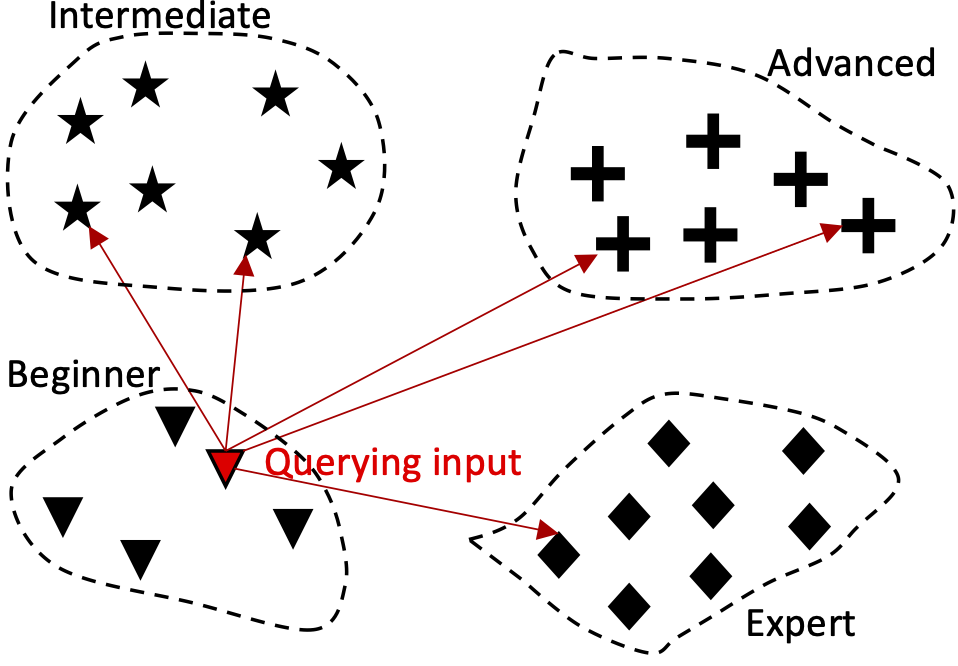}
	\caption{Counterfactual selection}
	\label{fig:example}
	\end{minipage}
%	\vspace{-2em}
\end{figure}

\sstitle{Motivating example}
\autoref{fig:example} illustrates a skill classification in an automated job application screening program~\cite{kirkpatrick2016battling}. A candidate was classified by a machine learning model into one of four expertise levels: beginner, intermediate, advanced, and expert. The applicant was classified as a beginner and rejected because an expert was needed. Typically, the company might provide an explanation like ``due to poor work experience'', but such explanations do not guide the applicant on how to improve their chances of future acceptance.

\sstitle{Limitations of existing works}
(1) Existing methods for counterfactual explanations often focus on binary outcomes (accept/reject) and unrealistic goals, like requiring a PhD, which are impractical for beginners~\cite{schleich2021geco}. Focusing on intermediate levels offers more actionable guidance. (2) Most approaches prioritise examples similar to the query, but diverse counterfactuals, such as bootcamps that improve skills in six months, are more beneficial. (3) Many generate synthetic examples~\cite{karimi2020survey}, which, while sound, can be unrealistic (e.g., an 18-year-old with a PhD). (4) Static, offline methods are unsuitable for streaming data, as they lack the speed and flexibility to generate real-time explanations, such as identifying trends from newly submitted applications.

\sstitle{Contributions}
To address these issues, we adopt a model-free approach to generate counterfactuals without requiring access to the original model, ensuring generalizability across domains (e.g., finance, marketing) and avoiding model-specific limitations~\cite{barocas2020hidden}. Unlike prior works providing a single counterfactual~\cite{vermeire2020explainable}, our method offers multiple alternatives, such as gaining skills, enhancing a portfolio, or adding referees in job applications. It is also free from model bias common in ML models~\cite{mehrabi2021survey}. To handle streaming data and deliver real-time feedback, we use a one-pass streaming algorithm.

In this paper, we propose a unified framework to select a subset of counterfactual examples for a query input, maximizing utility. Our method scales to any data stream size, requiring only a single pass without storing original items. \autoref{tab:functionality} compares our approach to state-of-the-art methods, evaluating key features: model access (model-free), example type (real or synthetic), on-the-fly computation (streaming), inclusion of multiple examples (multiplicity), and diversity of examples (diversity).

In summary, the contributions of this research include:

\begin{compactitem}
\item A first system to compute real counterfactual explanations in stream.
\item A mitigation of bias when selecting counterfactuals by incorporating multiple concepts of diversity: constraint-based, content-based, class-based, and sampling-based.
\item A subset selection problem that balances cardinality, similarity, and diversity for counterfactuals.
\item A generic streaming algorithm for counterfactual subset selection that processes the entire data stream in a single pass, achieving a \(1/5.585\) quality guarantee with \(O(\log k)\) complexity per item.
\item Experiments demonstrating our performance across multiple datasets and settings, showing that our approach computes explanations on the fly regardless of data size.
\end{compactitem}

\sstitle{Related work}
Recent interest in explanation methods spans interpretable models, rule-based, feature-based, and example-based explanations~\cite{dwivedi2023explainable}. Interpretable models (e.g., regressions, decision trees) allow inspection but require ML knowledge and may lack accuracy~\cite{mothilal2020explaining}. Feature-based methods (e.g., LIME, SHAP, LRP) rank feature importance but overlook explanation size, feature dependence, and multi-class predictions~\cite{karimi2020survey}. Traditional methods focus on \emph{what happened}, identifying contributing factors, but fail to provide actionable guidance on \emph{what to do next} for outcome improvement, e.g., ``poor experience'' as a job rejection reason~\cite{nguyen2015result,nguyen2017retaining,tam2019anomaly,nguyen2020monitoring,nguyen2019user,nguyen2020entity,nguyen2021structural}.

Multiple explanation methods~\cite{stepin2021survey} include global model explanations and instance-level predictions~\cite{stepin2021survey}. These can be model-specific (intrinsic) or model-agnostic (free). Model-specific methods are efficient and accurate but lack flexibility and generalizability~\cite{stepin2021survey}, while model-agnostic methods are versatile but may sacrifice performance~\cite{stepin2021survey,vermeire2020explainable}. Our work advances explanations independent of model evaluations~\cite{duong2022efficient,nguyen2020factcatch,hung2017answer,nguyen2017argument,ren2022prototype,nguyen2018if,toan2018diversifying}.

Counterfactual explanations~\cite{wachter2017counterfactual,verma2020counterfactual,byrne2019counterfactuals} describe how different circumstances could lead to alternative outcomes, offering actionable guidance, such as suggesting a rejected student improve their GRE score or gain research experience~\cite{mertes2022ganterfactual,schleich2021geco,BuiNN22,MaGMZL22,DuttaLMTM22}. While some methods provide multiple counterfactuals~\cite{schleich2021geco,BuiNN22}, many rely on synthetic examples, which can be unrealistic and less actionable~\cite{mertes2022ganterfactual,ustun2019actionable}. Humans prefer realistic counterfactuals~\cite{byrne2019counterfactuals}, and methods leveraging feature similarity offer some improvements~\cite{van2021interpretable,poyiadzi2020face,kanamori2020dace}, but they often fall short in feasibility and scalability. Our work addresses these gaps by selecting real counterfactuals, ensuring practical, achievable explanations~\cite{hollig2023semantic}. In contrast, many existing approaches generate synthetic counterfactuals, including adversarial examples~\cite{moore2019explaining}, perturbed data points~\cite{lash2017generalized}, or synthetic case-based examples~\cite{kenny2021generating,keane2020good}, highlighting the need for methods grounded in real data~\cite{zhao2021eires,huynh2021network,duong2022deep,nguyen2022model,nguyen2022detecting,trung2022learning,huynh2023efficient,thang2022nature}.

\section{Model and Problem Formulation}
\label{sec:model}

Consider a dataset \(D\) of \(n\) items, where each item has a single label \(l=1,2,\ldots,L\), and \(D_l\) denotes the set of items with label \(l\). The groups \(D_1, \ldots, D_L\) are disjoint, forming \(D = D_1 \cup \ldots \cup D_L\). The goal is to maximize a non-negative submodular function \(f: 2^D \times D \rightarrow \mathbb{R}_{\geq 0}\) for a querying item \(q \in D\). For simplicity, \(f(S)\) is used instead of \(f(S, q)\) since \(q\) is fixed. Here, \(f(.)\) represents a utility function over counterfactual explanations, as detailed in \autoref{sec:utility}.

\subsection{Counterfactual Diversity}

Offering a diverse set of counterfactuals provides opportunities to showcase data items different from the query, enabling users to explore varied explanations and reducing presentation bias~\cite{nguyen2019maximal,nguyen2021judo,nguyen2020factcatch,nguyen2022survey}. The goal is to ensure every item has an equal chance of selection, regardless of attributes or class~\cite{nguyen2024manipulating,nguyen2023validating,nguyen2019debunking}. We propose the following diversity constructs.

\sstitle{Constraint-based diversity} 
Bias can be mitigated with diversity constraints~\cite{russell2019efficient}. Following prior works~\cite{HalabiMNTT20}, we define a selection \(S \subseteq D\) as constraint-based diverse if it satisfies \(\alpha_l \leq |S \cap D_l| \leq \beta_l\), where \(\alpha_l, \beta_l \in \mathbb{Z}_{\geq 0}\). These bounds are often proportional to \(|D_l| / n\), ensuring balance with respect to label distribution. This definition generalizes bias countermeasures like the 80\%-rule and proportional representation~\cite{HalabiMNTT20}.

\sstitle{Content-based diversity}
This type of diversity aims to eliminate explanations containing many pairs of similar examples. Given a set of data items \(S \subseteq D\) and a similarity measure \(sim(.,.)\), an explanation based on content diversity is defined as
$ -\sum_{e \in S} \sum_{e' \in S} sim(e,e')$.

\sstitle{Sampling-based diversity}
This diversity type addresses sampling bias using a determinantal point process to minimise pairwise similarity. It is represented by the determinant of a distance kernel matrix, \(\text{det}(\mathbb{K}_S)\), where \(\mathbb{K}_{i,j} = \frac{1}{1+dist(e_i, e_j)}\) and \(dist(e_i, e_j) = 1-sim(e_i, e_j)\). To ensure well-defined determinants, small random perturbations are added to the diagonal elements during computation.

\sstitle{Clustering-based diversity}
This diversity metric leverages the clustering nature of items. The local coverage of a data item \(e\) over a label \(l\) is defined as \((1/2)^{1-sim(l,e)}\), where \(sim(l,e)\) is the maximum similarity between any item \(e'\) of label \(l\) and \(e\), and \(1/2\) is a decay factor. Based on local coverage, the global coverage of an item \(e\) is:
$
 cov(e) = \sum_{l \in L} (1/2)^{1-sim(l,e)}
 $.
 
The diversity of a set \(S\) of items is defined as a weighted sum of coverages:
\begin{equation}
 \sum_{e \in S} sim(e,q) cov(e)
 \end{equation}
The factor \(sim(e,q)\) avoids selecting items of low relevance but high diversity.

\subsection{Counterfactual Utility}
\label{sec:utility}

With these constructs of diversity, an utility measure for an optimal explanation can now be specified. We propose four measures: 
\begin{equation}
	\label{eq:content_explainability}
	\mbox{\textbf{Content-based utility:} } f_1(S) = \sum_{e \in S} sim(e,q) - \frac{\lambda_1}{|S|^2} \sum_{e}\sum_{e' \neq e} sim (e,e')
\end{equation} 
The parameter \(0 \leq \lambda_1 \leq 1\) balances similarity and diversity, with a larger \(\lambda_1\) emphasizing diversity. The first term is similar to the proximity in DiCE~\cite{mothilal2020explaining}, while the second term is inspired by~\cite{nguyen2015result}. This utility favors items that are similar to the query but also different from each other.
\begin{equation}
\mbox{\textbf{Sampling-based utility:} }	f_2(S) = \sum_{e \in S} sim(e,q) - \frac{\lambda_2}{|S|} det(\mathbb{K}_S)
\end{equation}
The parameter \(0 \leq \lambda_2 \leq 1\) balances similarity and diversity. \(\mathbb{K}_S\) is a pair-wise matrix of the distance metric for items in \(S\), where \(\mathbb{K}_{i,j} = \frac{1}{1+dist(e_i, e_j)}\) and \(dist(e_i, e_j) = 1 - sim(e_i, e_j)\). This approach is inspired by DPP~\cite{kulesza2012determinantal}, which have been used in diversified sampling.
\begin{equation}
	\label{eq:modality_explainability}
\mbox{\textbf{Clustering-based util:} }	f_3(S) = \sum_{e \in S} sim(e,q) + \frac{\lambda_3}{|S|} \sum_{e \in S} sim(e,q) cov_\alpha(e)
\end{equation}	
where $0 \leq \lambda_3 \leq 1$ is a parameter to balance the similarity and diversity in an explanation. A larger $\lambda_3$ implies the more emphasis on diversity. 
\begin{equation}
\label{eq:hybrid_explainability}
\mbox{\textbf{Hybrid utility:} } f(S) = f_1(S) + f_2(S) + f_3(S)
\end{equation}

%Before presenting the selection algorithms, there are two important properties of $f_1$, $f_2$, $f_3$, and $f$. Proofs are omitted due to space limit~\cite{jin2021unconstrained}.

%\begin{remark}[Monotonicity]
%	Let $D$ be a set of data items. For any subcollection $S \subseteq D$, we have $f_1(S) \leq f_1(S \cup \{e\})$ and $f_2(S) \leq f_2(S \cup \{e\})$ and $f_3(S) \leq f_3(S \cup \{e\})$ and $f(S) \leq f(S \cup \{e\})$ for all $e \in D \setminus S$. 
%\end{remark}

%\begin{remark}[Submodularity]
%Let \(D\) be a set of data items. For any item \(e \in D \setminus S\) in any subset \(S \subseteq D\), we have \(f_1(S \cup \{e\}) - f_1(S) \leq f_1(\{e\})\). Similar inequalities hold for \(f_2\), \(f_3\), and \(f\).
%\end{remark}

\subsection{Problem Statement}

For each label \(l\), lower and upper bounds (\(\alpha_l, \beta_l\)) define the number of labeled items \(l\) required, ensuring constraint-based diversity. With a global cardinality constraint \(k \in \mathbb{Z}_{\geq 0}\), the solution space \(\mathcal{S}\) is:
$ \mathcal{S} = \{S \subseteq D : |S| < k, \alpha_l \leq |S \cap D_l| \leq \beta_l, \forall l=1,\ldots,L\} $.
In \autoref{fig:example}, diversity constraints promote examples with varied features, while the cardinality constraint allows for examples across job levels, catering to user preferences.

\sstitle{Diversified submodular maximisation}
Maximizing \(f\) under cardinality and diversity constraints involves selecting \(S \subseteq D\) with \(S \in \mathcal{S}\) to maximize \(f(S)\). Let \(S^*\) denote the set that maximizes \(f\). We assume a feasible solution exists (\(\mathcal{S} \neq \emptyset\)), ensuring \(\sum_{l=1}^L \alpha_l \leq k\). Here, \(f(.)\) represents a utility function, not model calls as in prior works~\cite{mothilal2020explaining,schleich2021geco}.

\section{One-pass Explanation Algorithm}
\label{sec:approach}

This section presents our algorithm for counterfactual subset selection in streaming settings, including proof of its approximation guarantee, with the algorithm parsing the data only once. We start with foundational concepts.

\subsection{Foundation concepts}
\label{sec:concepts}

\sstitle{Marginal gain}
Given two sets \(S, S' \subseteq D\), the marginal gain of \(S\) with respect to \(S'\) is defined as:
$ f(S|S') = f(S \cup S') - f(S')$,
which quantifies the change in value when adding \(S\) to \(S'\). The function \(f\) is submodular if for any subset \(S \subseteq D\) and any item \(e \in D \setminus S\), we have:
$ f(\{e\} | S) \leq f(\{e\})$.
This means the marginal gain of any item is less than its value. For brevity, \(f(\{e\})\) is written as \(f(e)\), \(f(\{e\} | S)\) as \(f(e|S)\), \(S \setminus \{e\}\) as \(S \setminus e\), and \(S \cup \{e\}\) as \(S \cup e\).

\sstitle{Curvature}
A lower-bound for the marginal gain is derived by following a curvature of a function $f$:
$
	cv(f) = \min\{c: \forall S \subseteq D, \forall e \in D \setminus S, (1-c) f(e) \leq f(e|S)\}
$,
where $c \in (0,1]$. When $c=0$, $f$ is no longer submodular; rather, it is modular.

%\begin{remark}[Curvature]
%\label{remark:curvature}
%For a large enough $k$, we have $cv(f_1) \rightarrow 0$, $cv(f_2) \rightarrow 0$, $cv(f_3) \rightarrow 0$, and $cv(f) \rightarrow 0$.
%\end{remark}

\sstitle{Matroids}
A family of sets $\mathcal{F} \subseteq 2^D$ is called a matroid if it satisfies:
(1) $\mathcal{F} \neq \emptyset$;
(2) \emph{Downward-closedness:} if $F_1 \subseteq F_2$ and $F_2 \in \mathcal{F}$. then $F_1 \in \mathcal{F}$;
(3) \emph{Augmentation:} if $F_1,F_2 \in \mathcal{F}$ with $|F_1| < |F_2|$, then there exists $e \in F_2$ such that $F_1 \cup \{e\} \in \mathcal{F}$.

\sstitle{Extensibility}
A set \(S \subseteq D\) is called extensible if it can be a subset of some feasible solution \(S'\). For \(S\) to be extensible, it must satisfy the upper bounds \(|S| \leq k\) and \(|S \cap D_l| \leq \beta_l\) for all \(l=1..L\). This is necessary because \(S'\) is already feasible. However, \(S\) might violate the lower bounds (\(\alpha_l \leq |S \cap D_l|\) for all \(l\)), requiring the addition of at least \(\alpha_l - |S \cap D_l|\) items of each label \(l\), which could violate the global cardinality constraint \(k\).
Therefore, the necessary and sufficient conditions for a set \(S \subseteq D\) to be extensible are:
(E1) \(S \cap D_l \leq \beta_l\) for all \(l=1..L\) and
(E2) \(\sum_{l=1}^L \max(|S \cap D_l|, \alpha_l) \leq k\).

\subsection{Relaxed version}

Removing the lower-bound constraints \(|S \cap D_l| \geq \alpha_l\) from \(\mathcal{S}\) transforms it into a matroid, reducing the problem to submodular maximization under a matroid constraint. The goal is to select \(S \subseteq D\) with \(S \in \mathcal{F}\) to maximize \(f(S)\), where \(\mathcal{F}\) is a matroid, requiring a membership function to check \(S \in \mathcal{F}\)~\cite{chakrabarti2015submodular}. We propose a generic greedy streaming algorithm (\autoref{alg:matroid}), inspired by Chakrabarti et al.~\cite{chakrabarti2015submodular} and Huang et al.~\cite{huang2020approximability}, as a warm-up for our complete version. Each item \(e\) is processed once, computing the marginal gain (\autoref{line:marginal}) and checking feasibility (\autoref{line:matroid}) under constraints \(|S| \leq k\) and \(|S \cap D_l| \leq \beta_l\) for all \(l = 1..L\). If infeasible, the algorithm swaps \(e\) with a candidate \(e'\) based on minimal marginal gain (\autoref{line:minimum}) and the threshold \(\lambda\), ensuring \(e\)'s gain is at least \(1 + \lambda\) times \(e'\)'s gain (\autoref{line:heuristic}). The threshold \(\lambda\) governs quality guarantees.

\sstitle{Quality analysis}
In terms of solution quality, \autoref{alg:matroid} achieves an approximation ratio of \(\rho = 2(1+\lambda)^2 / \lambda - \lambda / (1+\lambda)^2 > 1\), with a minimum value of 7.75 at \(\lambda=1\). This means \(f(S) \geq 1/7.75 f(\text{OPT}) \geq 1/\rho f(\text{OPT})\), where \(\text{OPT}\) is the optimal solution. This is because \autoref{alg:matroid} is a special case of the problem SOTA for maximum submodular independence set (MSIS), and a solution for MSIS can be transformed into a solution for matroid submodular maximization in constant time~\cite{chakrabarti2015submodular}. Although these utility functions are well-designed, the approximation ratio can still be further improved.

%\begin{remark}[Quality guarantee]
%\label{remark:quality}
%	\autoref{alg:matroid} can produce a solution with an approximation ratio of $5.585$ in terms of $f_1$, $f_2$, $f_3$, and $f$ for a large enough $k$. That is, $f_1(S) \geq 1/5.585 f_1(OPT)$, $f_2(S) \geq 1/5.585 f_2(OPT)$, $f_3(S) \geq 1/5.585 f_3(OPT)$, and $f(S) \geq 1/5.585 f(OPT)$.
%\end{remark}

%\begin{proof}[Sketch]
%\autoref{alg:matroid} is a special case of the SOTA algorithm for the maximum weighted matching (MWM) problem, offering an approximation ratio of \(5.585/(1-\text{cv}(f))\) with \(\lambda = 0.717\). It can be transformed into a solution for matroid submodular maximization in constant time~\cite{chakrabarti2015submodular}. According to \autoref{remark:curvature}, which states that \(\text{cv}(f) \approx 0\) for a large enough \(k\), we have \(5.585/(1-\text{cv}(f)) \approx 5.585\).
%\end{proof}
%\vspace{-2em}

\begin{algorithm}[!h]
 \caption{Matroid Submodular Maximisation}
 \label{alg:matroid}
\renewcommand{\algorithmicrequire}{\textbf{Input:}}
\renewcommand{\algorithmicensure}{\textbf{Output:}}
 \begin{algorithmic}[1]
\REQUIRE{ $D$, $f$, $\mathcal{F}$}
\ENSURE{ A feasible solution $S \in \mathcal{F}$}
 \STATE $S = \emptyset$ \;
 \FOR{every arriving item $e$}
 	\STATE $w(e) = f(e | S)$ \; \label{line:marginal}
 	\IF{$S \cup e \in \mathcal{F}$}\label{line:matroid}
 	\STATE	$S = S \cup e$
 	\ELSE
 		\STATE $\Delta = \{e' \in S: S \cup e \setminus e' \in \mathcal{F} \}$ \; \label{line:candidate}
 		$e' = \argmin_{e' \in \Delta} w(e')$ \; \label{line:minimum}
 		\IF{$w(e) \geq (1+\lambda) w(e')$}\label{line:heuristic}
 			\STATE $S = S \cup e \setminus e'$
 		\ENDIF
 	\ENDIF
 \ENDFOR
 \RETURN{S}
  \end{algorithmic}
\end{algorithm}

%\vspace{-2em}
To select the best \(\lambda\) for a small \(k\), check if \(5.585/(1-\text{cv}(f)) > 7.75\). If true, set \(\lambda\) to \(0.717\); otherwise, set it to \(1\).

\sstitle{Complexity analysis}
The complexity of involves runtime, utility calls, matroid queries, and space. \autoref{alg:matroid} makes two utility calls per item to compute \(f(e|S)\). Runtime is dominated by matroid queries, candidate set construction, and minimum selection, implemented in \(O(k)\) time and matroid queries, where \(k\) is the rank of matroid \(\mathcal{F}\) (\(|S| \leq k\)). Query runtimes depend on the specific matroid. Space complexity is \(O(k)\), as discarded item weights need not be stored.

%\begin{remark}[$k$-uniform matroid]
%\label{remark:matroid}
%For special matroids \(\mathcal{F}\), \autoref{alg:matroid} can be optimized. Consider the special case where \(\mathcal{F}\) is the \(k\)-uniform matroid (\(S \in \mathcal{F} \Leftrightarrow |S| \leq k\)), keeping only the cardinality constraint. In this case, the matroid membership query (\autoref{line:matroid}) takes \(O(1)\) time, and the candidate set construction (\autoref{line:candidate}) becomes \(\Delta = S\). The runtime is then dominated by finding the item \(e' \in S\) with the lowest historical marginal gain \(w\). By maintaining a priority queue \(W\) containing \(S\) sorted by \(w\), this can be done in \(O(\log k)\) time.
%\end{remark}

\subsection{The complete algorithm}

\sstitle{Intuition}
This section outlines the intuition behind our algorithm and introduces the complete algorithm line by line. For practical use, \autoref{alg:matroid} must function in a black-box manner, though its solution may violate lower-bound constraints. To address this, ``backup'' items can be gathered in parallel to \autoref{alg:matroid}. As the function is monotone submodular, adding these items preserves the approximation guarantee but may violate the cardinality constraint \( |S| \leq k \). Not every set satisfying upper-bound constraints is extensible to a feasible solution, but we show that extendable subsets of \(D\) form a matroid, enabling efficient solution.

%\begin{remark}[Extensibility matroid]
%Let $\hat{S} \subseteq 2^D$ contain all extensible subsets of $D$. Then $\hat{S}$ is a matroid. This can be derived from the results of \cite{HalabiMNTT20}. 
%\end{remark}

\sstitle{Putting it all together}
\autoref{alg:explanation} presents a complex algorithm for diversified submodular maximization. It uses \autoref{alg:matroid}, a streaming algorithm for submodular maximization under a matroid constraint, as a black box on the extensibility matroid \(\hat{S}\) to construct an extensible set \(S_1\) that approximately maximizes \(f\). In parallel, it collects a preserve set \(P_l\) of size \(|P_l| = \alpha_l\) for every label \(l\). Finally, the solution \(S_1\) is augmented into a feasible solution \(S\) by adding items from \(P_l\) to \(S_1\) to satisfy any lower bound constraints where \(|S_1 \cap D_l| < \alpha_l\).
%\vspace{-2em}

\begin{algorithm}[H]
\renewcommand{\algorithmicrequire}{\textbf{Input:}}
\renewcommand{\algorithmicensure}{\textbf{Output:}}
\caption{One-pass explanation algorithm}
 \label{alg:explanation}
 \begin{algorithmic}[1]
\REQUIRE $D$, $f$, $\mathcal{S}$
\ENSURE A feasible solution $S \in \mathcal{S}$
 \STATE {$S_1 = \emptyset$;
 $P_l = \emptyset, \forall l=1..L$;}
 \FOR{every arriving item $e$ of label $l$}
 \STATE	{process $e$ with \autoref{alg:matroid} ($D$, $f$, $\hat{S}$)}
 	\IF{$|P_l| < \alpha_l$}
 		\STATE $P_l = P_l \cup e$
 	\ENDIF
 \ENDFOR
 \STATE $S_1$ = solution of \autoref{alg:matroid} ($D$, $f$, $\hat{S}$) \;
 \STATE $S$ = $S_1$ augmented with items in sets $P_l$\;
 \RETURN{$S$}
 \end{algorithmic}
\end{algorithm}

%\vspace{-2.5em}
\sstitle{Constraint checking}
To run \autoref{alg:matroid} efficiently on the extensibility matroid \(\hat{S}\), we need to check if adding a single item \(e\) to an extensible set \(S\) maintains extensibility in \(O(1)\) time. This is done by maintaining the counts \(c_l = |S \cap D_l|\) for each label \(l\) in \(S\), and the sum \(C = \sum_{l=1}^L \max(c_l, \alpha_l)\). By definition, \(S\) is extensible if \(c_l \leq \beta_l\) for each \(l\) and \(C \leq k\).
Initially, these variables are set as:
 (i) \(c_l = 0\) for all \(l = 1..L\),
 (ii) \(C = \sum_{l=1}^L \alpha_l\).
When a potential extension \(e\) with label \(l\) is considered, the algorithm checks extensibility if \(c_l < \alpha_l \vee \alpha_l \leq c_l < \beta_l \wedge C < k\). When \(S\) is augmented with \(e\):
(i) \(c_l\) is incremented by 1,
(ii) If \(c_l > \alpha_l\), \(C\) is incremented by 1.

\sstitle{Quality guarantee}
\autoref{alg:explanation} returns a feasible solution belonging to \(\mathcal{S}\) with the same approximation ratio as \autoref{alg:matroid}. 
First, the feasibility of \(S\) is ensured because \(S_1\) is extensible, satisfying the lower-bound constraints without violating the cardinality constraint \(k\). 
Second, since \autoref{alg:matroid} is a \(1/\rho\)-approximation algorithm, it returns a solution \(S_1\) with a value \(f(S_1)\) that is at least \(1/\rho\) times that of the best extensible set. 
Moreover, \(\mathcal{S} \subseteq \hat{S}\) by definition, and adding items to \(S_1\) does not decrease \(f(.)\) since \(f\) is monotone. Thus, the optimal solution in \(\hat{S}\) has a value for \(f(.)\) greater than or equal to that of the optimal solution in \(\mathcal{S}\). 
Therefore, \autoref{alg:explanation} maintains the same approximation ratio as \autoref{alg:matroid}.

%\begin{remark}[Offline comparison]
%\label{remark:offline}
%	There does not exist an offline algorithm with an approximation ratio higher than $1-1/e = 0.632$ and a polynomial number of utility calls.
%\end{remark}
%
%\begin{proof}[Sketch]
%Our problem generalizes the constrained maximization~\cite{chakrabarti2015submodular}, which has shown that given an upper bound of \(\rho \geq e/(e-1)\), a \(1/\rho\)-approximation algorithm must make a super-polynomial \#calls to the utility function.
%\end{proof}

%According to \autoref{remark:offline} and \autoref{remark:quality}, no algorithm can produce a solution that is $3.53\times$ better than \autoref{alg:explanation} without exceeding a polynomial number of utility calls for a sufficiently large \(k\).

\sstitle{Runtime and memory}
We can further optimize our algorithm to be extremely efficient, using only \(O(\log k)\) time per item. The space complexity remains \(O(k)\) as \autoref{alg:explanation} uses extra memory for \(|\bigcup_l P_l| = \sum_l \alpha_l \leq k\). The algorithm still makes two utility calls per item via \autoref{alg:matroid}.
Extending the idea in \autoref{remark:matroid} to \(\mathcal{F}\) being an extensibility matroid, we maintain a data structure for the query, i.e., \(c_l = |S \cap D_l|\) and \(C=\sum_{l=1}^L \max(c_l, \alpha_l)\). To find the minimal \(w(e')\) among \(e' \in \Delta\), which are items \(e' \in S\) such that \(S \cup e \setminus e' \in \mathcal{F}\), we check whether a label \(l'\) is good. The logic is:
(i) If \(c_l = \beta_l\), then only \(l\) is good.
(ii) If \(C < k\) or \(c_l < \alpha_l\), then every color is good.
(iii) Otherwise, good labels are \(l\) and those \(l'\) with \(c_{l'} > \alpha_{l'}\).
To quickly find the minimum-weight item with a good label in \(S\), we maintain several priority queues:
(i) \(W\), containing \(S\) sorted by \(w\).
(ii) \(W_l\), for each label \(l\), containing items in \(S \cap D_c\) sorted by \(w\).
(iii) \(W'\), containing labels with \(c_{l'} > \alpha_{l'}\), sorted by \(\min_{e' \in S \cap D_{l'}} w(e')\).
This data structure can be maintained in \(O(\log k)\) time per item while implementing the above logic.

\section{Empirical Evaluation}
\label{sec:exp}

\sstitle{Real datasets}
We evaluated our method using three representative datasets: 
\begin{compactitem}
	\item \emph{Income}~\cite{celis2018fair}: 
This dataset contains 45,222 records from the Census database with 14 attributes, including age, race, gender, education, income, occupation, and marital status. Race was selected as the label, ensuring its distribution in \(S\) matched the dataset~\cite{HalabiMNTT20}. We set \(\alpha_l = \lfloor 0.9 \frac{|D_l|}{|D|} k \rfloor\) and \(\beta_l = \lceil 1.1 \frac{|D_l|}{|D|} k \rceil\), varying \(k\).	

	\item \emph{Customer}~\cite{moro2014data}: 
This dataset includes phone call records from a bank's marketing campaign, with features such as client age, gender, account balance, call date, and duration. Age was chosen as the sensitive feature, dividing records into six groups: [0,29] (5273), [30,39] (18,089), [40,49] (11,655), [50,59] (8410), [60,69] (1230), and [70+] (554). Diversity constraints ensured each group formed 10-20\% of the subset~\cite{HalabiMNTT20}.

	\item \emph{Lending Credit~\cite{davenport2015lending}:}%
This dataset includes five years of loan data from LendingClub, a peer-to-peer lending platform. After preprocessing~\cite{mothilal2020explaining}, we retained 7 features: employment years, annual income, open credit accounts, credit history, home ownership, purpose, and state. The sensitive feature was the loan grade assigned by LendingClub, with diversity constraints ensuring a similar grade distribution.

\end{compactitem}

\sstitle{Synthetic data}
We used Synthea~\cite{chen2019validity}, a synthetic data generator, to create realistic clinical data for over one million synthetic residents of Massachusetts. This population reflects Massachusetts' demographics, disease burdens, vaccinations, and social determinants. Disease modules served as sensitive labels, with diversity constraints ensuring similar disease distributions.

\sstitle{Evaluation metrics}
Given their inherent subjectivity, counterfactual examples are challenging to evaluate. While not intended to replace behavioral experiments, we formulated three metrics: 
\begin{compactitem}
	\item \emph{Transport cost:}
This distance-based cost measure indexes the usefulness of counterfactual examples. Users prefer counterfactual examples similar to the query example, minimizing the effort required to achieve a desired class outcome in another feature.%
\begin{equation}
cost = \frac{1}{|S|} \sum_{e \in S} d_{con}(e,q) + d_{cat}(e,q)
\end{equation}
where $d_{con}(e,q)$ is the l1-distance across continuous features between $e$ and $q$. $d_{cat}(e,q)$ is the l1-distance between $e$ and $q$ across their categorical features. 

\item \emph{No. constraint violation:} 
This metric counts the number of diversity constraint violations during streaming, reflecting the plausibility of outcomes. Smaller values indicate reduced bias. As our work addresses presentation bias rather than algorithmic bias~\cite{mehrabi2021survey}, fairness metrics like equalized odds and statistical parity are not applicable, as they target entire datasets rather than individual explanations~\cite{mehrabi2021survey}.

\item \emph{Runtime:} measures the total running time of the explanation process.
\end{compactitem}

\sstitle{Baselines}
The following baselines were proposed:

\begin{compactitem}
	\item \emph{Offline:} 
This baseline processes the entire data stream with multiple passes, solving the subset selection problem. In each pass, it selects the item \(e\) with maximal marginal gain \(f(e|S)\) such that \(S \cup e\) remains extensible~\cite{HalabiMNTT20}. The algorithm achieves a \(1/2\)-approximation with a runtime of \(O(|D| k)\) and \(O(|D| k)\) oracle calls.

	\item \emph{kNN:} 
This non-streaming baseline processes the entire data stream using nearest-neighbor search~\cite{keane2020good}. Given disjoint groups \(D_1, \ldots, D_L\) with the same label (\autoref{sec:model}), the algorithm rotates between groups, selecting the most similar item and removing it for the next rotation, skipping groups if label cardinality constraints are violated. The kNN time complexity is \(\mathcal{O}(k|D|)\), reducible to \(\mathcal{O}(k\log|D|)\) using structures like R-Tree, though tree construction requires \(\mathcal{O}(|D|\log|D|)\).

	\item \emph{Random:} This baseline is designed to maintain a random feasible solution. The process is similar to \autoref{alg:explanation}, but the swapping mechanism is performed randomly instead of using the greedy threshold $\lambda$, i.e., every arriving time $e'$ has a $0.5$ chance of being swapped with a random item in $S$.

	\item \emph{Relaxed:} This is the relaxed version of our algorithm, which removes the lower-bound constraints.

	\item \emph{NoConstraint:} This is the state-of-the-art streaming submodular maximisation algorithm without any diversity constraints~\cite{badanidiyuru2014streaming}.
	\end{compactitem}
	
	\noindent
	State-of-the-art methods for counterfactual explanation~\cite{mertes2022ganterfactual,schleich2021geco,BuiNN22} focus on generating synthetic examples, while our method provides real examples~\cite{DuttaLMTM22,van2021interpretable,LeyBW22,PawelczykDHKL23}, making these approaches orthogonal to ours~\cite{guidotti2022counterfactual,MaGMZL22}. Our work emphasizes efficient computation with a quality approximation guarantee. Including these methods~\cite{guidotti2021ensemble,keane2020good,kenny2021generating,moore2019explaining,lash2017generalized}  would be unfair and potentially misleading.

\sstitle{Environment}
All experiments were on an Intel i7 system (3.4GHz, 12GB RAM). The results are averaged over 10 runs (variances reported if appropriate).

\subsection{End-to-end Comparisons}
\label{sec:exp_end2end}

\begin{figure}[!h]
%\vspace{-3em}
\centering
\subfloat[Income dataset.\label{fig:effort_d1}]{%
     \includegraphics[width=.3\linewidth]{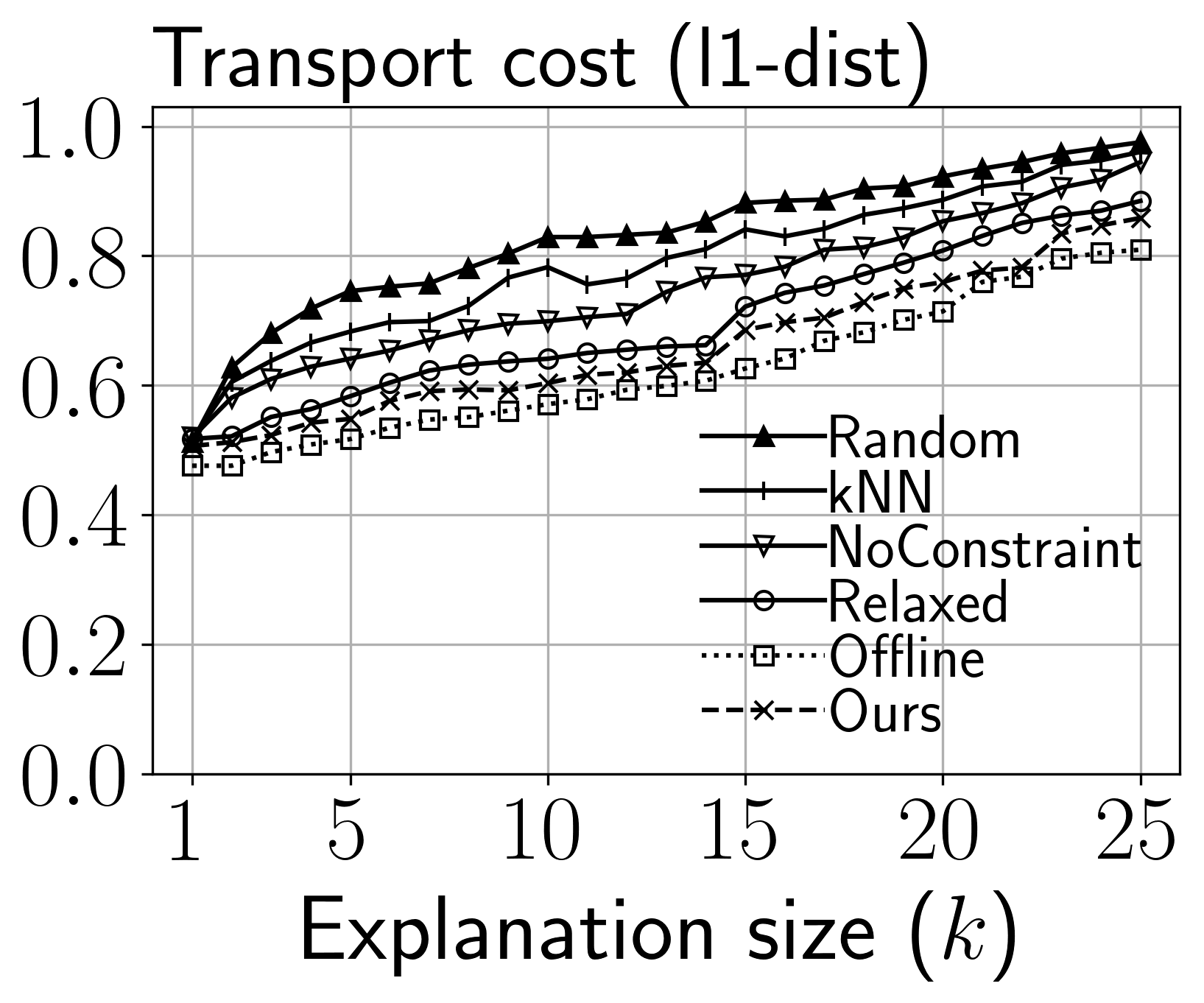}
    }
    \subfloat[Customer dataset.\label{fig:effort_d2}]{%
     \includegraphics[width=.3\linewidth]{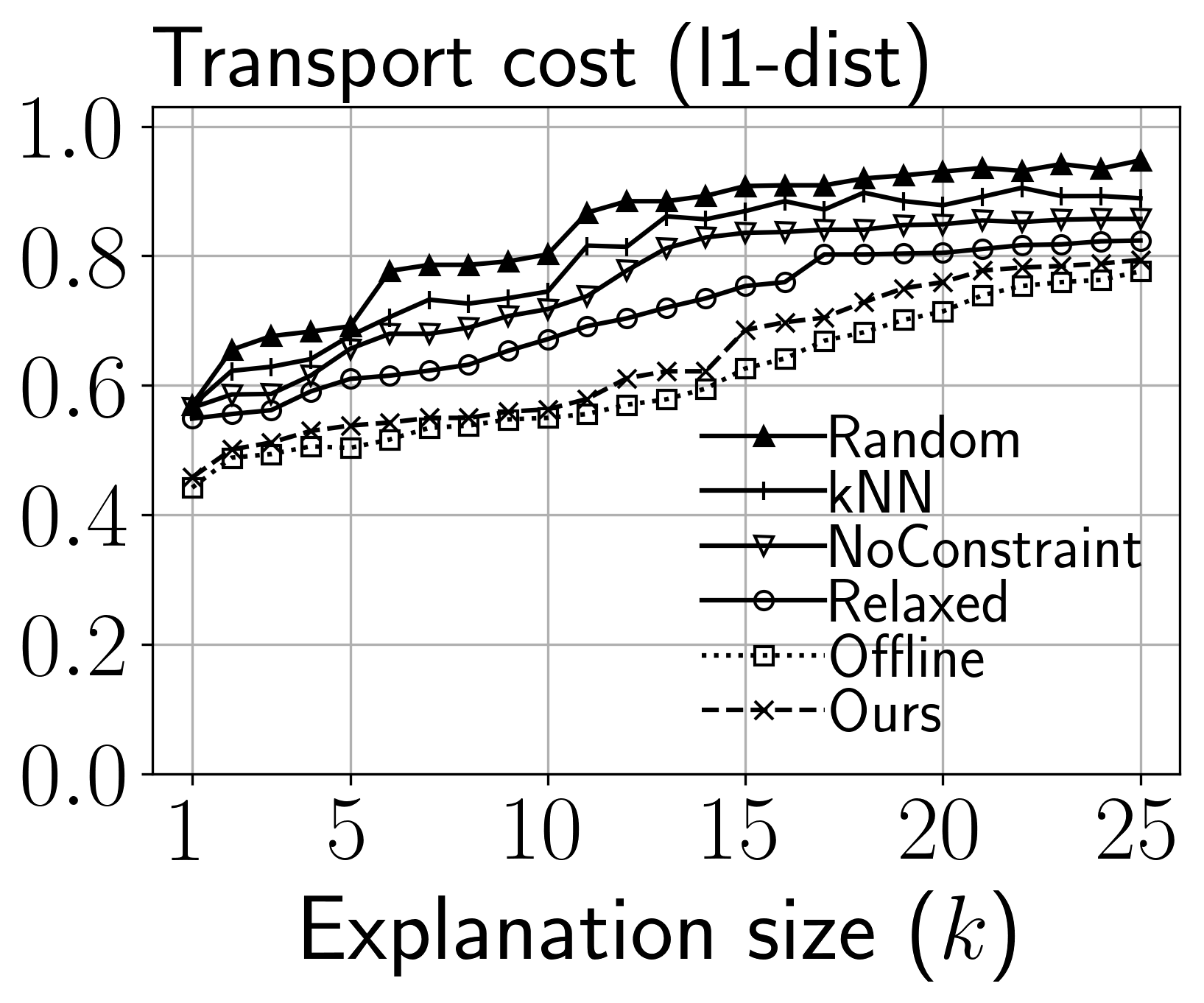}
    }
    \subfloat[Credit dataset.\label{fig:effort_d3}]{%
     \includegraphics[width=.3\linewidth]{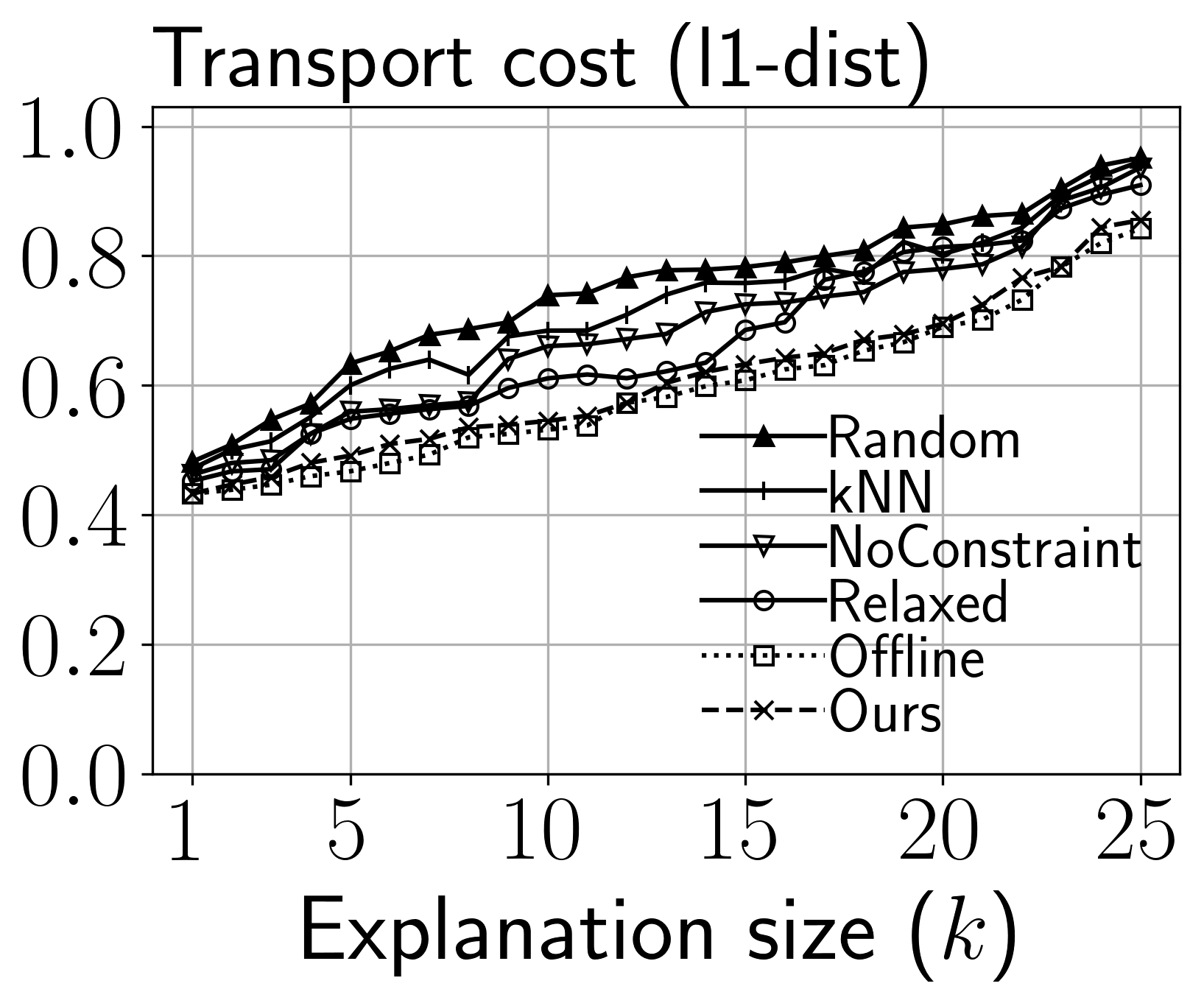}
    }
      \vspace{-1em}
        \caption{Effectiveness for real-world data.}
        \label{fig:effort_real}
%        \vspace{-2em}
 \end{figure}

\sstitle{Transport cost}
We compared our approach to baselines for generating user-friendly counterfactual examples using real datasets. \autoref{fig:effort_real} presents the results, with the X-axis showing the explanation size ($k$) and the Y-axis the transport cost as l1-distance. Our method outperformed streaming baselines (\emph{Random}, \emph{Relaxed}, \emph{NoConstraint}) and the non-streaming baseline (\emph{kNN}), matching the offline algorithm in some cases. For example, in the Customer dataset ($k=10$), our approach reduced transport costs by up to 33.18\%, meaning users exerted only two-thirds the effort to follow recommendations.

\begin{figure}[!h]
%\vspace{-3em}
\centering
\subfloat[Income dataset.\label{fig:violation_d1}]{%
     \includegraphics[width=.3\linewidth]{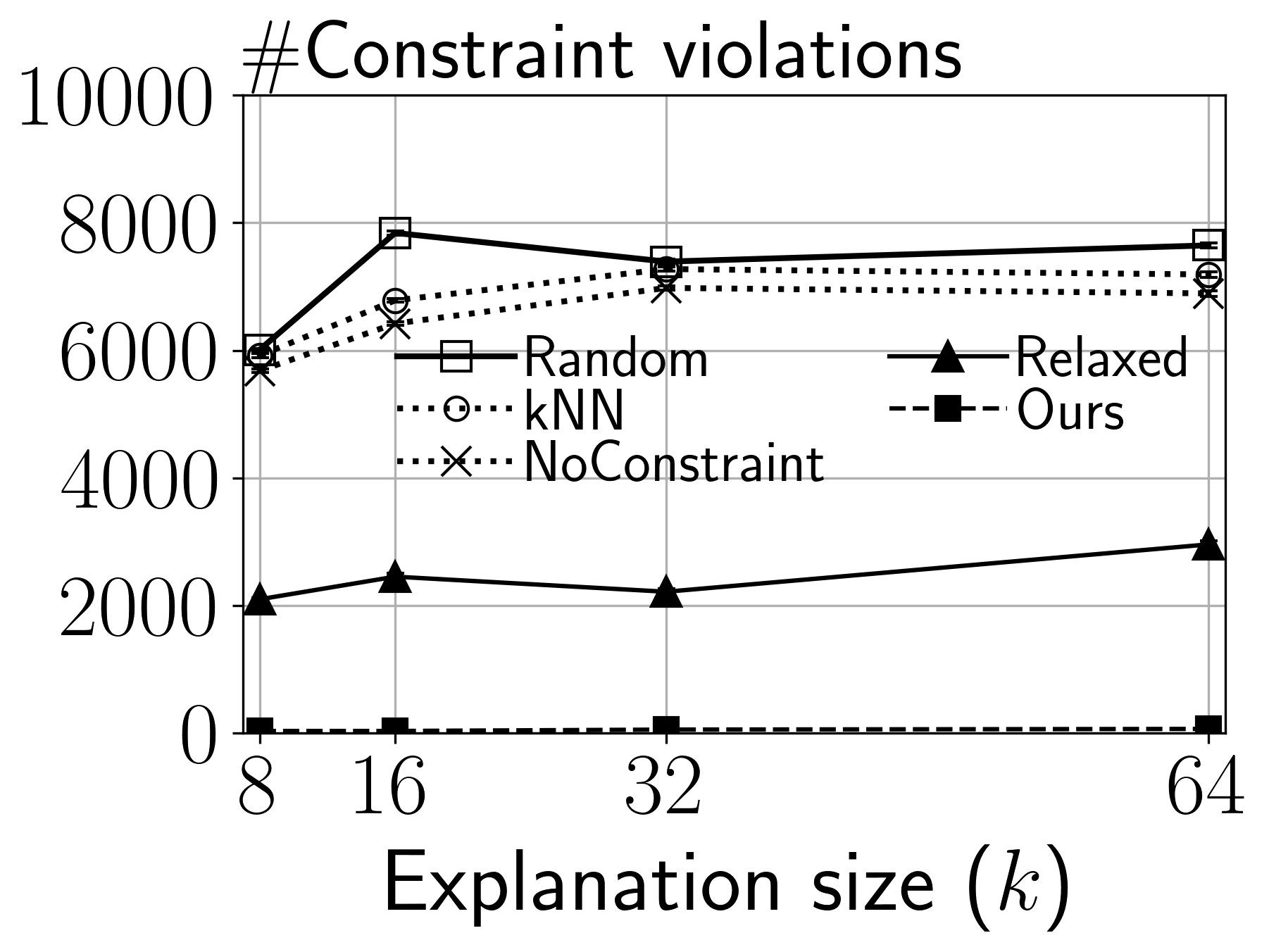}
    }
    \subfloat[Customer dataset.\label{fig:violation_d2}]{%
     \includegraphics[width=.3\linewidth]{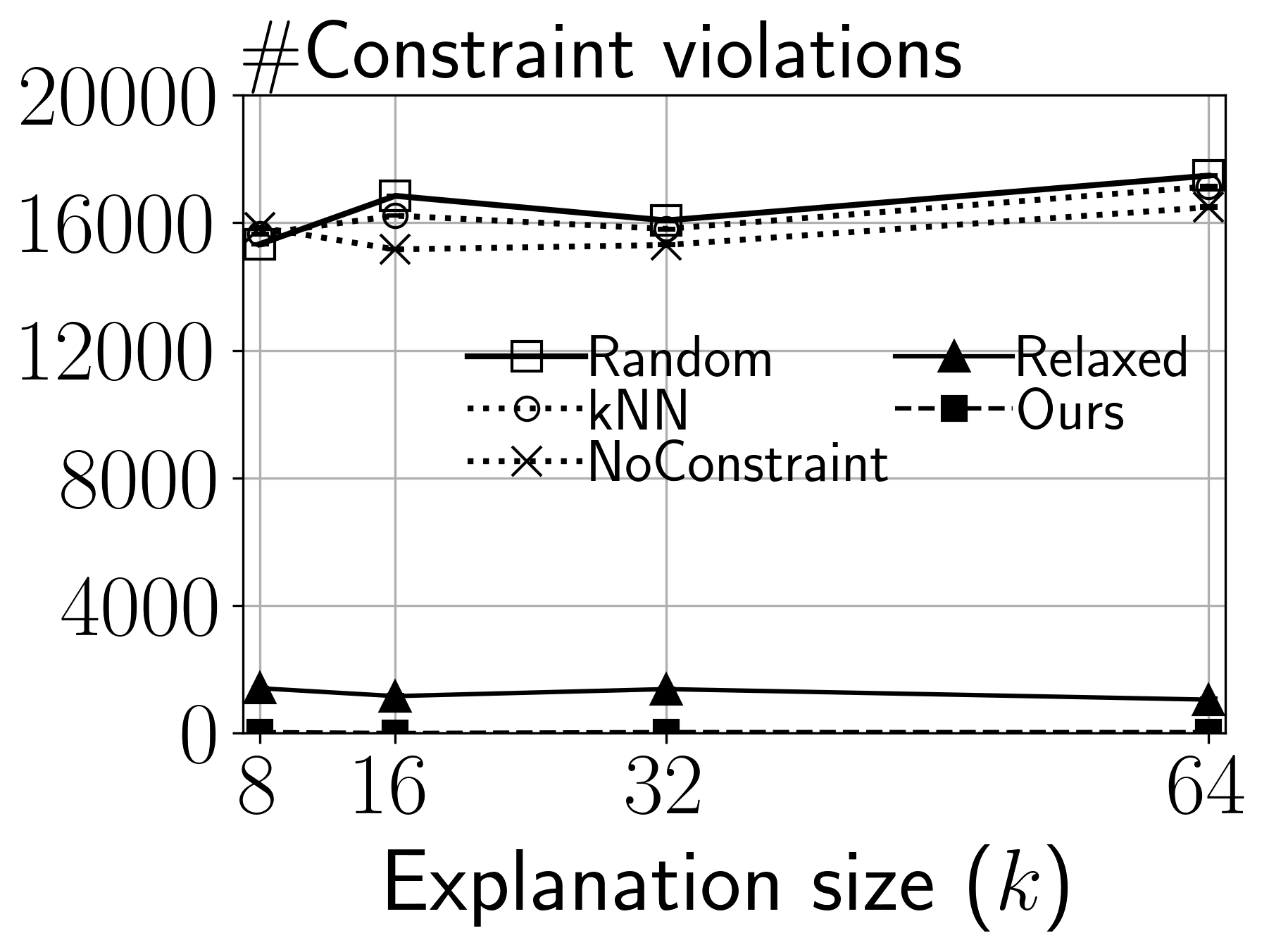}
    }
    \subfloat[Credit dataset.\label{fig:violation_d3}]{%
     \includegraphics[width=.3\linewidth]{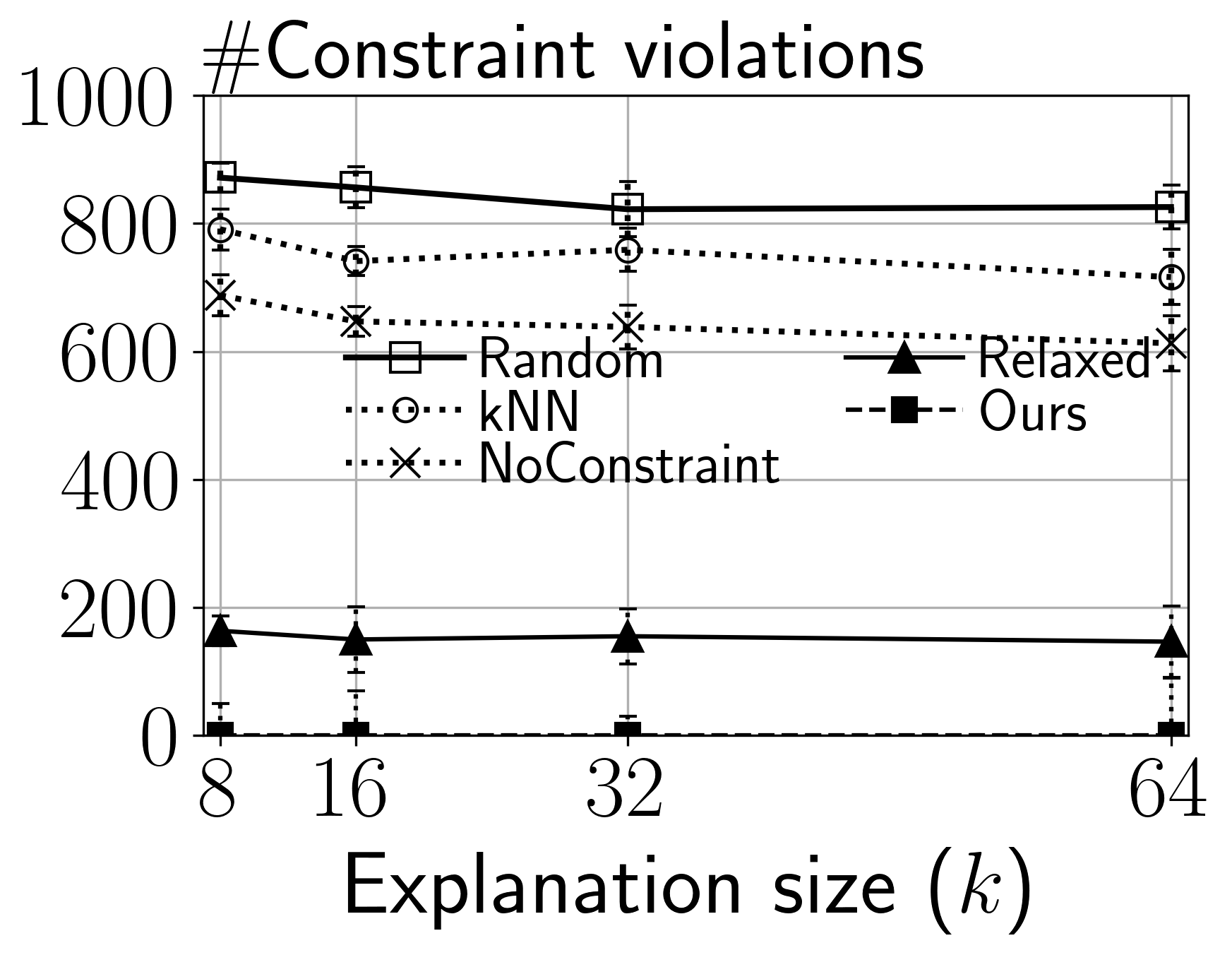}
    }
      \vspace{-1em}
        \caption{Constraint Violations in explanations.}
        \label{fig:violation}
%        \vspace{-2em}
 \end{figure}

 \sstitle{Constraint violations}
 We evaluated the plausibility of counterfactual explanations by comparing constraint violations across real datasets. \autoref{fig:violation} shows results, with the X-axis indicating the number of counterfactual examples ($k$) and the Y-axis the number of constraint violations. Our approach, like the \emph{offline} algorithm, had no violations (not shown for brevity). The \emph{Relaxed} version had some violations but far fewer than baselines such as \emph{Random}, \emph{kNN}, and \emph{NoConstraint}, which reached violation levels proportional to dataset size due to ignoring diversity constraints.

\begin{figure}[!h]
%\vspace{-3em}
\centering
\subfloat[Income.\label{fig:time_d1}]{%
     \includegraphics[width=.23\linewidth]{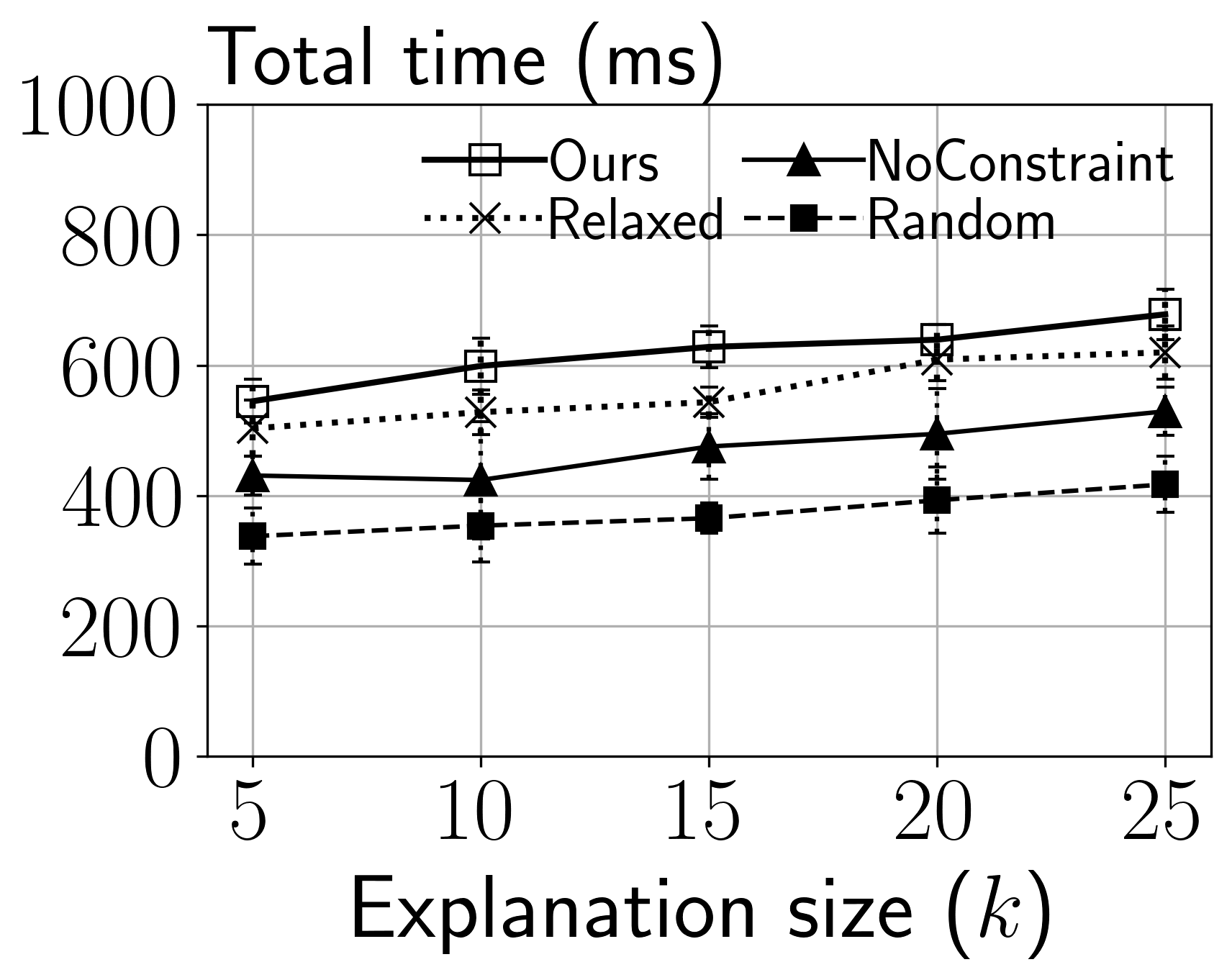}
    }
    \subfloat[Customer.\label{fig:time_d2}]{%
     \includegraphics[width=.23\linewidth]{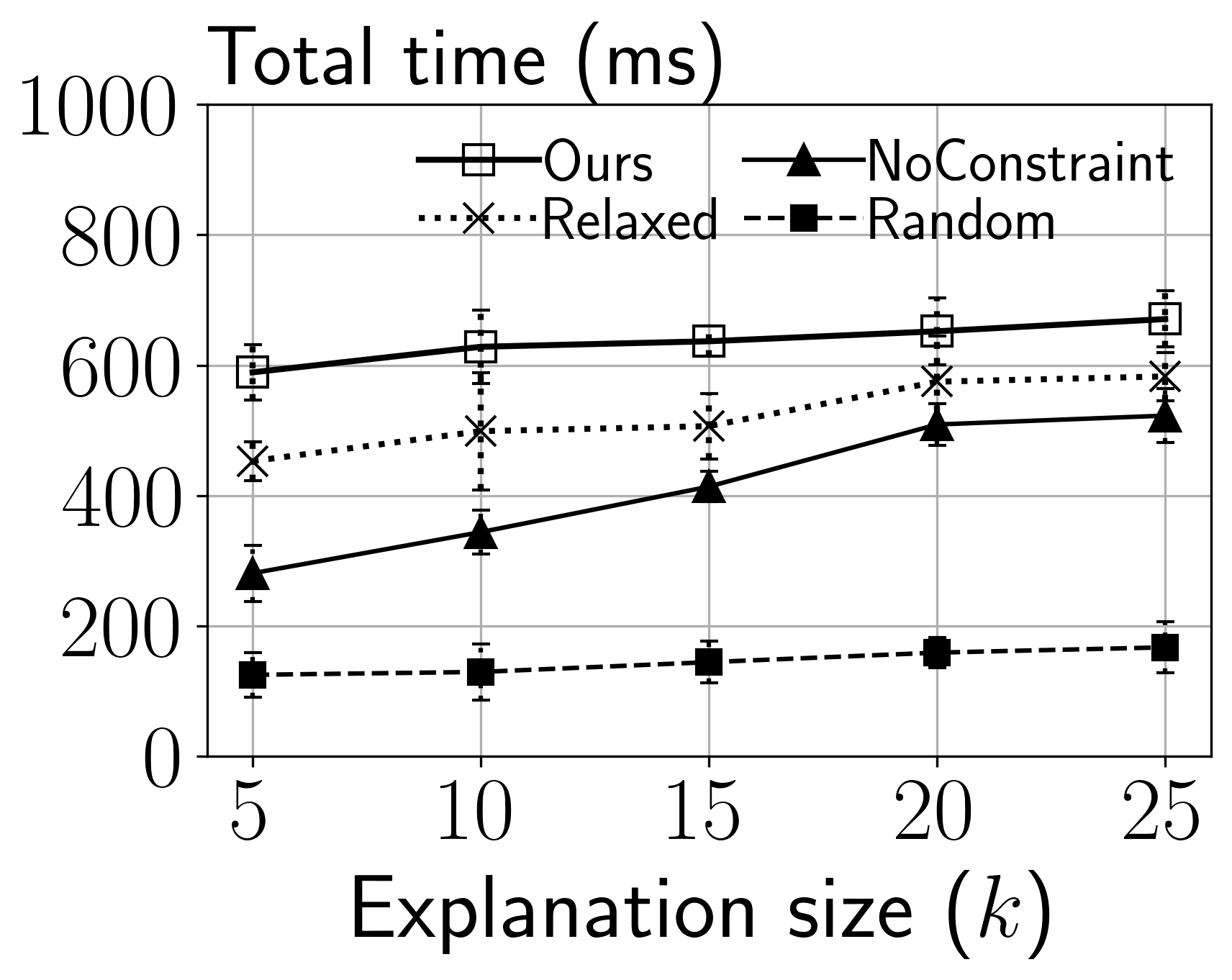}
    }
    \subfloat[Credit.\label{fig:time_d3}]{%
     \includegraphics[width=.23\linewidth]{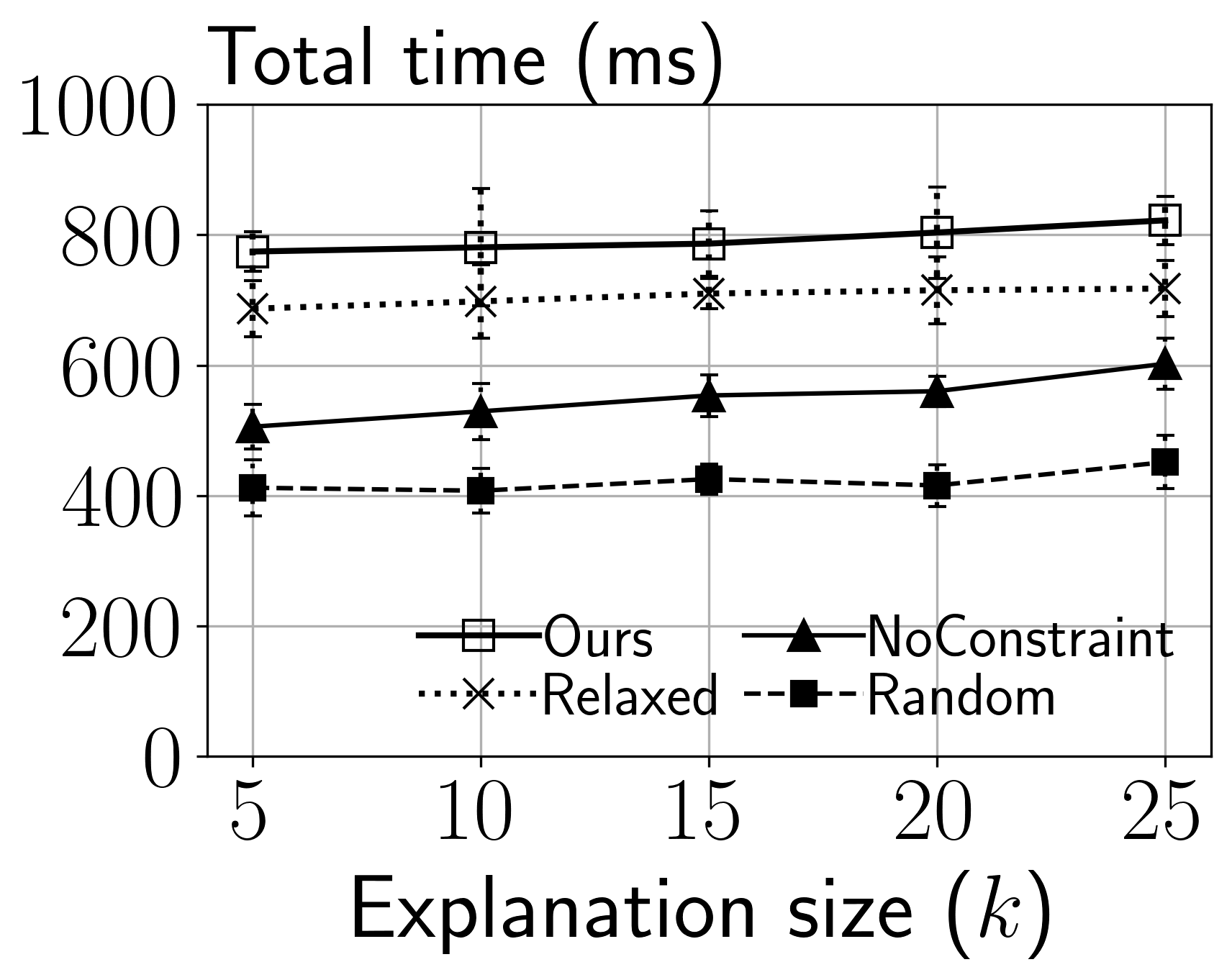}
    }
    \subfloat[Massachusetts.\label{fig:time_d4}]{%
     \includegraphics[width=.23\linewidth]{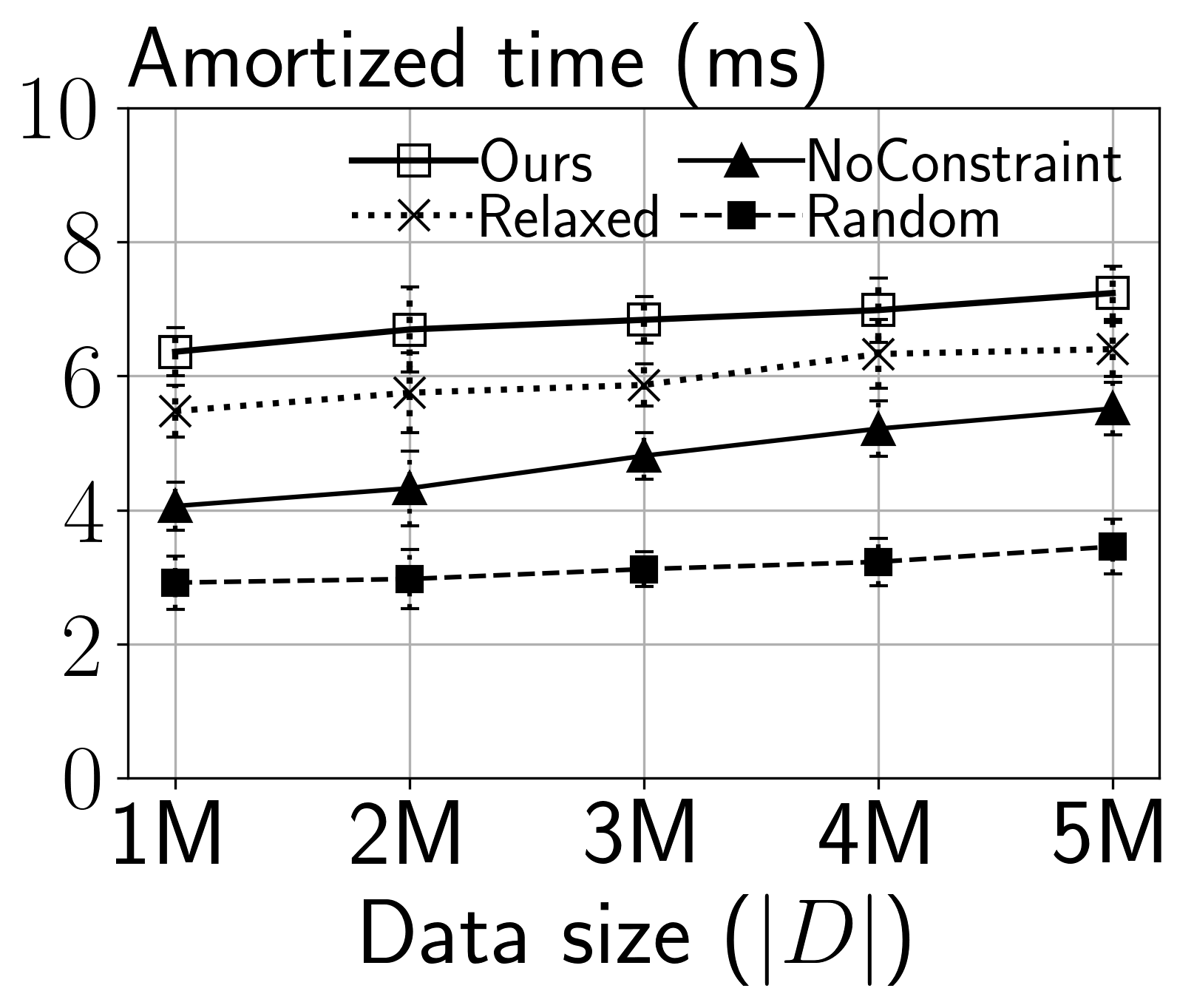}
    }
      \vspace{-1em}
        \caption{Runtime on different datasets.}
        \label{fig:time}
%	\vspace{-3em}
 \end{figure}

\sstitle{Runtime}
We evaluated the total runtime of our approach against baselines. Non-streaming methods, such as \emph{offline} and \emph{kNN}, had significantly longer runtimes and are excluded for clarity. \autoref{fig:time_d1}, \ref{fig:time_d2}, and \ref{fig:time_d3} present results, with the X-axis showing the explanation size ($k$) and the Y-axis the runtime. We varied $k$ up to 25 due to cognitive load limits. While our approach was slower than other baselines, maintaining additional structures and constraints, its runtime remained under 1 second, suitable for streaming. The framework computes explanations in $\mathcal{O}(\log k)$ per item (amortized), yielding $\mathcal{O}(n \log k)$ for data size $|D|=n$. In \autoref{fig:time_d4}, using Massachusetts synthetic data (sizes 1M-5M, $k=1000$), our method demonstrated the most efficient amortized time, $\mathcal{O}(\log k)$ per new item, regardless of data size.

\subsection{Ablation Study}
\label{sec:exp_ablation}

We evaluated our design choices by comparing variations of our approach:
(i) \emph{w/o sketch} -- no caching of utility values or constraint checks;
(ii) \emph{w/o lower-bound} -- ignores lower-bound constraints $\alpha_l$;
(iii) \emph{w/o upper-bound} -- uses only a $k$-uniform matroid instead of the extensibility matroid;
(iv) \emph{greedy threshold} -- fixes the threshold $\lambda$.
\autoref{tab:ablation} summarizes the results. Without sketches, runtime increased due to frequent utility evaluations and checks. Relaxing constraints improved utility and transport cost but increased violations. Fixing $\lambda$ reduced example quality. Notably, $\lambda=1$ matches the SOTA streaming algorithm for submodular maximization under fairness constraints~\cite{HalabiMNTT20}.

\begin{table}[!h]
%\vspace{-3em}
 \caption{Importance of each model component.}
  \label{tab:ablation}
  \centering
  \resizebox{0.8\linewidth}{!}{
  \begin{tabular}{lrrrr}
    \toprule
    full model & $\Delta$-Utility & $\Delta$-Transport Cost & $\Delta$-Constraint violations & $\Delta$-Running time  \\
    \midrule
      w/o sketch & +0\% & +0\% & +0\% & +893.25\% \\
     \midrule
      w/o lower-bound & +5.81\% & -2.75\% & +21.76\% & -12.41\% \\
      w/o upper-bound & +4.69\%  & -1.58\% & +33.54\% & -18.96\% \\
      \midrule
      $\lambda=0.5$ & -5.07\% & +12.78\% & +0\% & +0\%\\
      $\lambda=1$ (\cite{HalabiMNTT20}) & -5.91\% & +11.32\% & +0\% & +0\%\\
      $\lambda=2$ & -9.72\% & +19.64\% & +0\% & +0\% \\
    \bottomrule
  \end{tabular}
  }
%  \vspace{-3em}
\end{table}

\subsection{Robustness Analysis}
\label{sec:exp_robustness}

\begin{figure}[!h]
%\vspace{-3em}
\centering
\subfloat[Income dataset.\label{fig:robust_income}]{%
     \includegraphics[width=.31\linewidth]{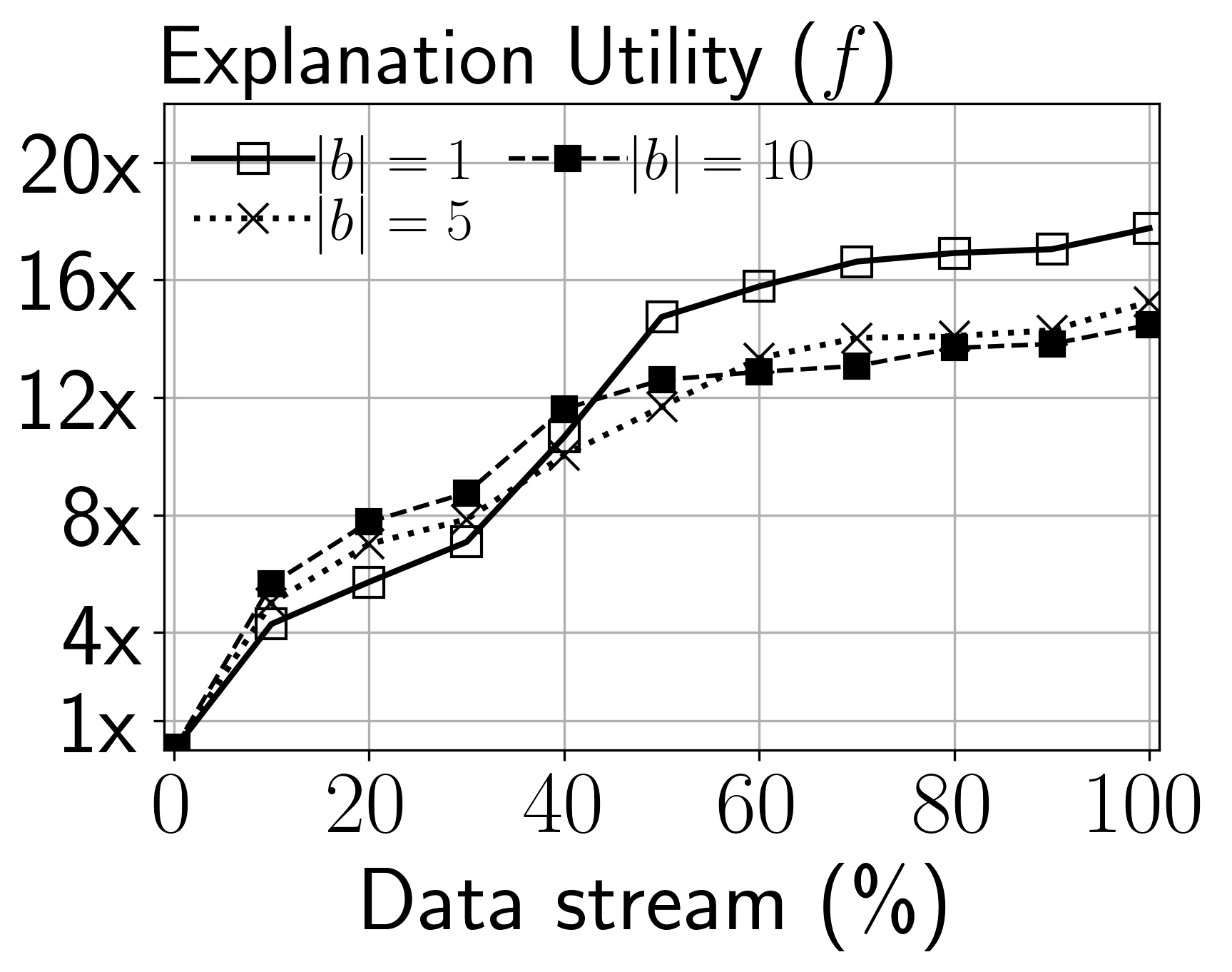}
    }
    \subfloat[Customer dataset.\label{fig:robust_customer}]{%
     \includegraphics[width=.31\linewidth]{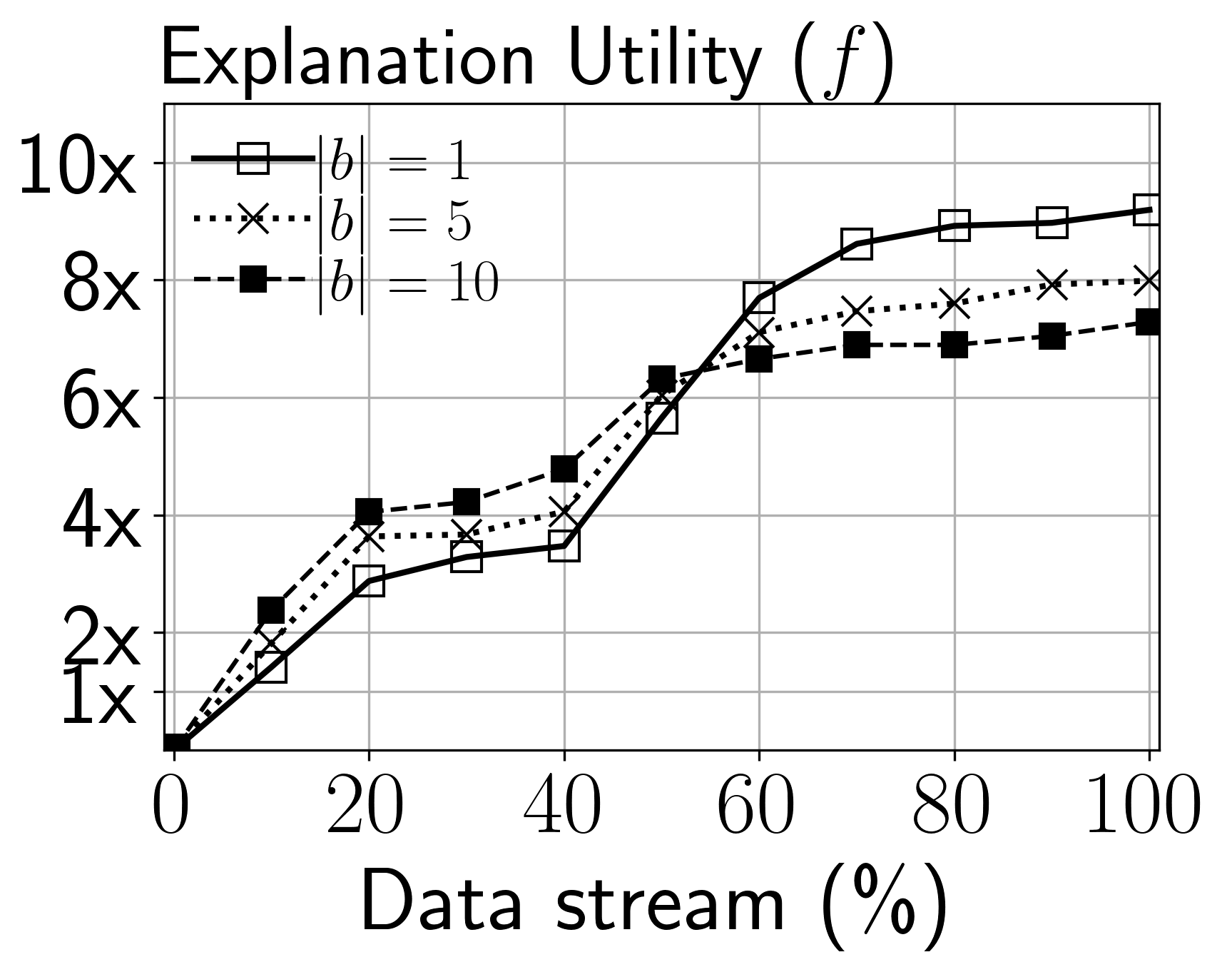}
    }
    \subfloat[Credit dataset.\label{fig:robust_credit}]{%
     \includegraphics[width=.31\linewidth]{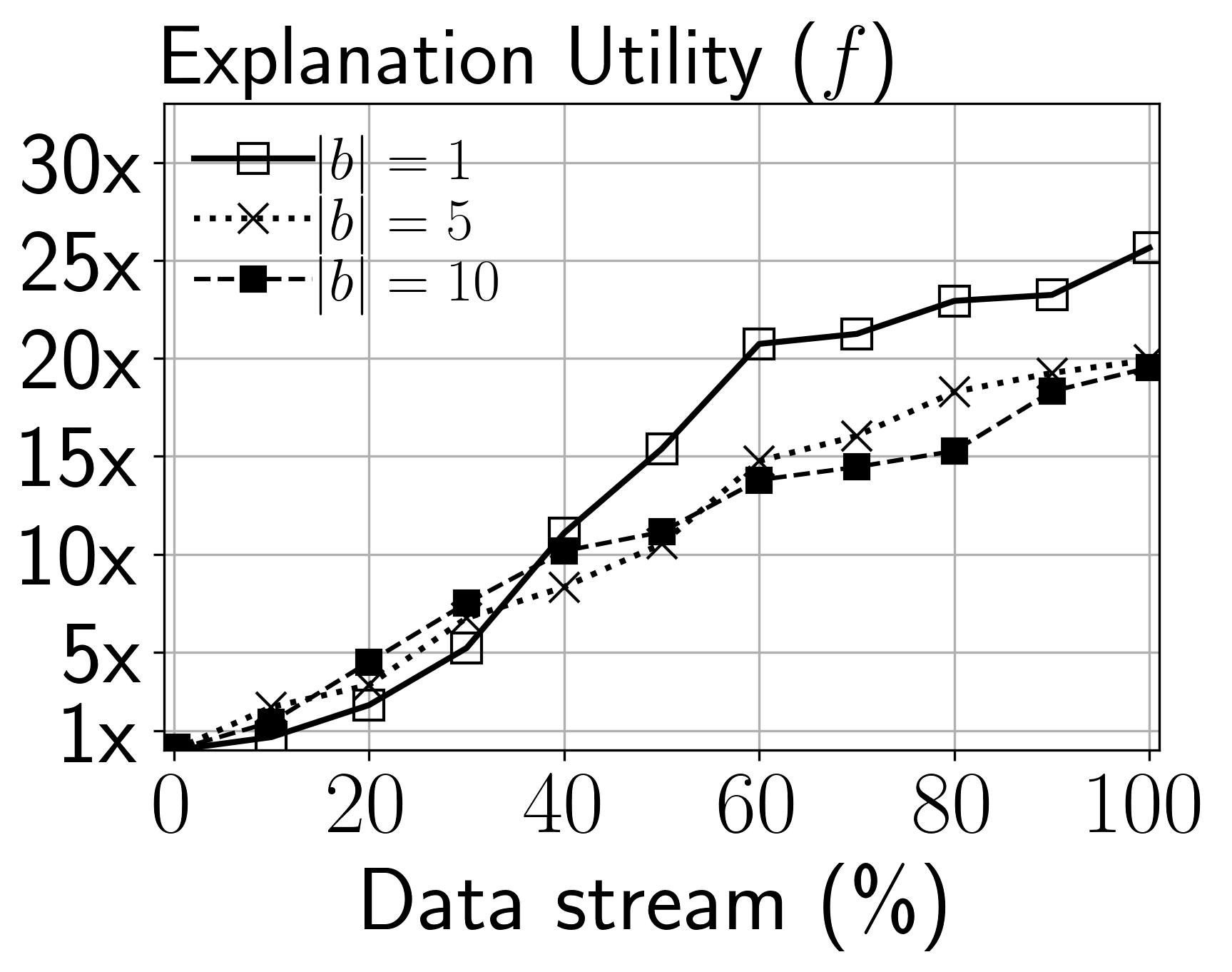}
    }
      \vspace{-1em}
        \caption{Effects of streaming data on explanation.}
        \label{fig:stream}
%        \vspace{-1em}
 \end{figure}

\sstitle{Sliding windows}
We evaluated the effect of streaming progress on utility by comparing our approach with its ``batch'' version using sliding windows of size $|b|$, where $b = \{e_1, \ldots, e_{|b|}\}$ is fed to the algorithm. The algorithm selects a subset $S'$ of $S \cup b$ (size $k$) with the highest $f(S')$, retaining only $S'$ per iteration. \autoref{fig:stream} shows results, with the X-axis as the percentage of data processed and the Y-axis as utility change. Utility increases with data stream progress, confirming the algorithm's soundness. Larger windows ($|b| = 5, 10$) initially increase utility faster (up to 40-60\%) as subsets are retained, but the original approach ($|b| = 1$) eventually outperforms due to lower regret from discarding key items as similarity distributions emerge.

\begin{figure}[!h]
%	\vspace{-3em}
	\centering
	 \begin{minipage}{.4\linewidth}
    \centering
    \captionsetup{type=table}
      \captionof{table}{Explanation utility in adverse conditions.}
  \label{tab:adaptive}
  \resizebox{0.99\linewidth}{!}{
  \begin{tabular}{clccc}
    \toprule
     Feature dist. & Label dist. & Ours & NoConstraint & Random \\
    \midrule
    \multirow{2}{*}{normal} & normal & \colorbox{green}{+0.00\%} & \colorbox{lime}{-19.85\%} & \colorbox{orange}{-30.69\%}  \\
     & skewed & \colorbox{green}{-1.63\%} & \colorbox{lime}{-24.77\%} & \colorbox{orange}{-39.47\%}  \\
    \midrule
    \multirow{2}{*}{skewed} & normal & \colorbox{yellow}{-3.72\%} & \colorbox{lime}{-31.26\%} & \colorbox{red}{-51.93\%}  \\
     & skewed & \colorbox{yellow}{-5.14\%} & \colorbox{lime}{-36.12\%} & \colorbox{red}{-62.64\%}  \\
    \bottomrule
  \end{tabular}
  }
      \end{minipage}
      \begin{minipage}{0.59\linewidth}
		\centering
		\includegraphics[width=1.0\linewidth]{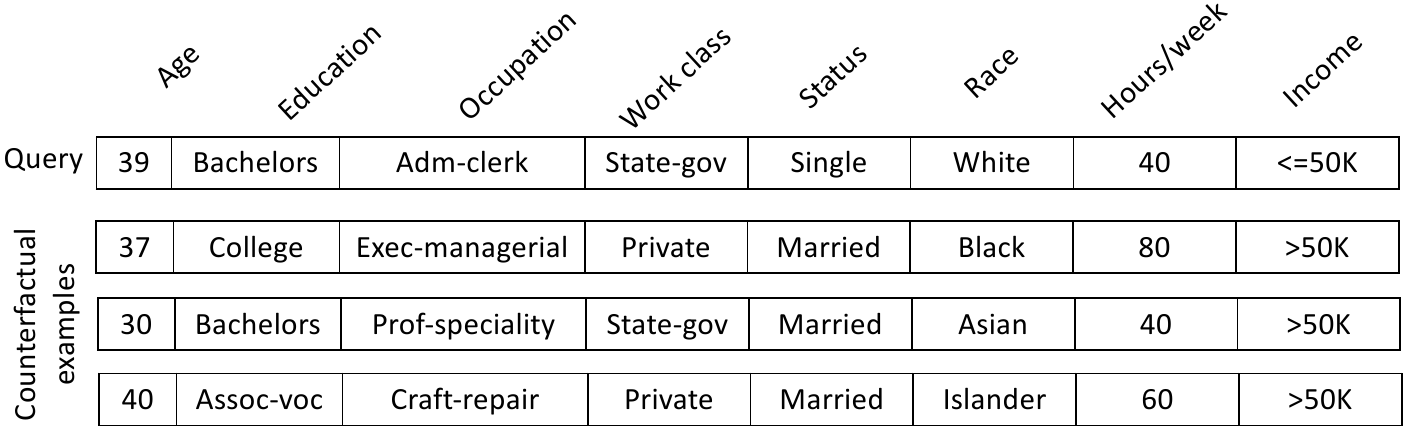}
	\caption{Examples of counterfactuals.}
	\label{fig:exp_quality}
	\end{minipage}
%	\vspace{-2.5em}
\end{figure}

\sstitle{Concept drifts}
We evaluated the robustness of streaming algorithms to concept drift using synthetic data with varied label distributions: (i) \emph{normal} -- future label distributions similar to the past, and (ii) \emph{skewed} -- future label distributions significantly different. Similarly, item feature distributions were \emph{normal} or \emph{skewed}. \autoref{tab:adaptive} shows the utility difference between baselines and our algorithm. The relaxed version performed the same as the complete algorithm and is omitted for brevity. Our algorithm remained robust under adversarial conditions, with a maximum degradation of 5.14\%, compared to up to 62.64\% for baselines in the worst case (skewed feature and class distributions). This robustness stems from maintaining diversity in label constraints and item features.

\sstitle{Case Study}
\autoref{fig:exp_quality} illustrates the querying input and counterfactuals ($k=3$) returned by our approach on the Income dataset. The results show a diverse yet relevant set of counterfactual examples. While our work focuses on efficiently computing these explanations, evaluating their usefulness requires user studies, which is beyond the scope of this paper.

\section{Conclusion}
\label{sec:conclusion}

We explored counterfactual explanations without relying on predictive models, minimizing bias and ensuring broader applicability. By prioritizing both label diversity and feature variability, our approach offers users more actionable pathways to influence outcomes. We formulated the problem as a subset selection task, balancing similarity and diversity while maintaining computational efficiency. To achieve this, we developed a scalable streaming algorithm capable of handling large, continuously arriving datasets without storing the full data history. Empirical results on real and synthetic datasets indicate our method reduces user effort by up to 33\%, enhances utility by 57.5\%, and improves data stream efficiency by a factor of 25. Future work can extend the framework to multi-label and fairness-aware counterfactual selection, ensuring broader applicability while mitigating biases in decision-making~\cite{nguyen2024instruction,nguyen2024manipulating,nguyen2025privacy,pham2024dual,nguyen2024multi,nguyen2024handling}. Enhancing adversarial robustness and energy-efficient processing would make the approach more reliable in real-time, resource-constrained environments such as edge devices~\cite{nguyen2023poisoning,nguyen2023example,nguyen2014reconciling,nguyen2015smart,thang2015evaluation,nguyen2015tag,hung2019handling}. Additionally, exploring human-in-the-loop feedback, graph-based selection, and distributed processing could improve interpretability, scalability, and adaptability across diverse domains like healthcare, finance, and criminal justice~\cite{yang2024pdc,sakong2024higher,huynh2024fast,huynh2025certified,nguyen2023isomorphic,nguyen2024portable,ren2024comprehensive}.

%\bibliographystyle{elsarticle-num-names}
%\bibliography{../ref,../ref_h}

\end{document}